%% file: main.tex
\documentclass[conference, 10 pt, conference]{ieee/ieeeconf}  

\IEEEoverridecommandlockouts                             
\usepackage{graphics} 
\usepackage{graphicx}
\usepackage{epsfig} 
\usepackage{mathptmx} 
\usepackage{times} 
\usepackage{amsmath} 
\usepackage{amssymb}  
\usepackage[table]{xcolor}
\usepackage{color}
\usepackage[linesnumbered,ruled,vlined]{algorithm2e}
\usepackage{algorithmic}
\usepackage{subfigure}
\usepackage{multirow}
\usepackage{float}
\usepackage{algorithm2e}
\usepackage{booktabs}
\usepackage[utf8x]{inputenc}
\usepackage{wrapfig}
\usepackage{threeparttable}
\usepackage{boldline}
\usepackage{bbm}
\usepackage{bm}
\usepackage{interval}
\usepackage{microtype}
\let\labelindent\relax
\usepackage{enumitem}
\usepackage{dsfont}

\usepackage[binary-units,group-separator={,}]{siunitx}
\usepackage{etoolbox}
\newcommand{\ubold}{\fontseries{b}\selectfont}
\robustify\ubold
\newrobustcmd*{\bftabnum}{%
  \bfseries
  \sisetup{output-decimal-marker={\textmd{.}}}%
}

\usepackage{array}
    \newcolumntype{P}[1]{>{\centering\arraybackslash}p{#1}}
    \newcolumntype{M}[1]{>{\centering\arraybackslash}m{#1}}
\robustify\bfseries

\usepackage{caption}
\newcommand{\maximize}{%
  \mathopen{}\operatorname*{maximize}%
}

\newcommand{\specialcell}[2][c]{%
  \begin{tabular}[#1]{@{}c@{}}#2\end{tabular}}

\DeclareMathAlphabet{\mathcal}{OMS}{cmsy}{m}{n}
\renewcommand{\baselinestretch}{1}

\SetKwComment{Comment}{$\triangleright$\ }{}

\usepackage{xcolor}

\title{\LARGE \bf
	Learning-based Initialization of Trajectory Optimization
	
	for Path-following Problems of Redundant Manipulators
}

\author{Minsung Yoon\textsuperscript{\textdagger}, Mincheul Kang\textsuperscript{\textdagger}, Daehyung Park\textsuperscript{\textdagger\textdaggerdbl} and Sung-Eui Yoon\textsuperscript{\textdagger\textdaggerdbl}
\thanks{
    {\textsuperscript{\textdagger}}M. Yoon, M. Kang, D. Park, and S. Yoon are with the School of Computing, Korea Advanced Institute of Science and Technology, South Korea;{\tt\small \{minsung.yoon, mincheul.kang, daehyung\}@kaist.ac.kr, and sungeui@kaist.edu}. 
    {\textsuperscript{\textdaggerdbl}}D. Park and S. Yoon are co-corresponding authors.}}

\newcommand{\Skip}[1]{}
\renewcommand{\paragraph}[1]{{\bf {#1}}}  

\def\HiLi{\leavevmode\rlap{\hbox to 
		\hsize{\color{yellow!50}\leaders\hrule height .8\baselineskip depth .5ex\hfill}}}

\begin{document}
    \maketitle
    \thispagestyle{empty}
    \pagestyle{empty}
    
    \newtheorem{thm}{Theorem}
    \newtheorem{lem}[thm]{Lemma}
    \newtheorem{col}[thm]{Corollary}
    
    \input{abstract}
    \input{1}
    \input{3_1}
    \input{3_2}
    \input{4}
    \input{5}

    \section*{Acknowledgment}
    This work was supported in part by the Institute of Information \& communications Technology Planning \& Evaluation (IITP) and the National Research Foundation of Korea (NRF) grants, which are funded by the Korea government (MSIT), under IITP-2015-0-00199, No. 2021R1A4A1032582, and No. NRF-2021R1A4A3032834.

    {
        \small
        \bibliographystyle{ieee/ieee}
        \bibliography{./ref}
    }
    
\end{document}

%% file: abstract.tex
\begin{abstract}
Trajectory optimization (TO) is an efficient tool to generate a redundant manipulator's joint trajectory following a 6-dimensional Cartesian path.
The optimization performance largely depends on the quality of initial trajectories. 
However, the selection of a high-quality initial trajectory is non-trivial and requires a considerable time budget due to the extremely large space of the solution trajectories and the lack of prior knowledge about task constraints in configuration space.
To alleviate the issue, 
we present a learning-based initial trajectory generation method that generates high-quality initial trajectories in a short time budget by adopting example-guided reinforcement learning.
In addition, we suggest a null-space projected imitation reward to consider null-space constraints by efficiently learning kinematically feasible motion captured in expert demonstrations.
Our statistical evaluation in simulation shows the improved optimality, efficiency, and applicability of TO when we plug in our method's output, compared with three other baselines.
We also show the performance improvement and feasibility via real-world experiments with a seven-degree-of-freedom manipulator.

\end{abstract}

%% file: 1.tex
\section{Introduction}
\label{sec:1}

Tracing a 6-dimensional Cartesian path (i.e., target path) with an end-effector of a manipulator is an underpinning capability for an autonomous robot to perform everyday tasks such as welding, writing, and painting.
However, the manipulation control, especially for a redundant manipulator with an infinite number of inverse kinematics (IK) solutions given a single end-effector pose, is challenging in that a joint trajectory should minimize path-following objectives while satisfying a wide variety of constraints such as joint limits in range and velocity, singularity, and collision avoidance.

As a solution, researchers have typically used resolved motion rate control schemes that iteratively find a local solution via differential IK~\cite{bruyninckx2001open, beeson2015trac, siciliano1990kinematic, rakita2018relaxedik}.
However, the myopic aspect and numerical instability at singularities derive the resulted trajectory to the boundary of feasible configuration space.
Alternatively, a global-search method constructs a discrete configuration graph consisting of IK solutions along the target path and searches for the most kinematically feasible joint trajectory~\cite{rakita2019stampede, praveena2019user, holladay2019minimizing}.
This method is asymptotically optimal but slow and computationally expensive for the redundant manipulator to approximate a feasible configuration manifold due to numerous IK and constraint calculations.

\begin{figure}[t]
    \centering 
    \includegraphics[width=\columnwidth]{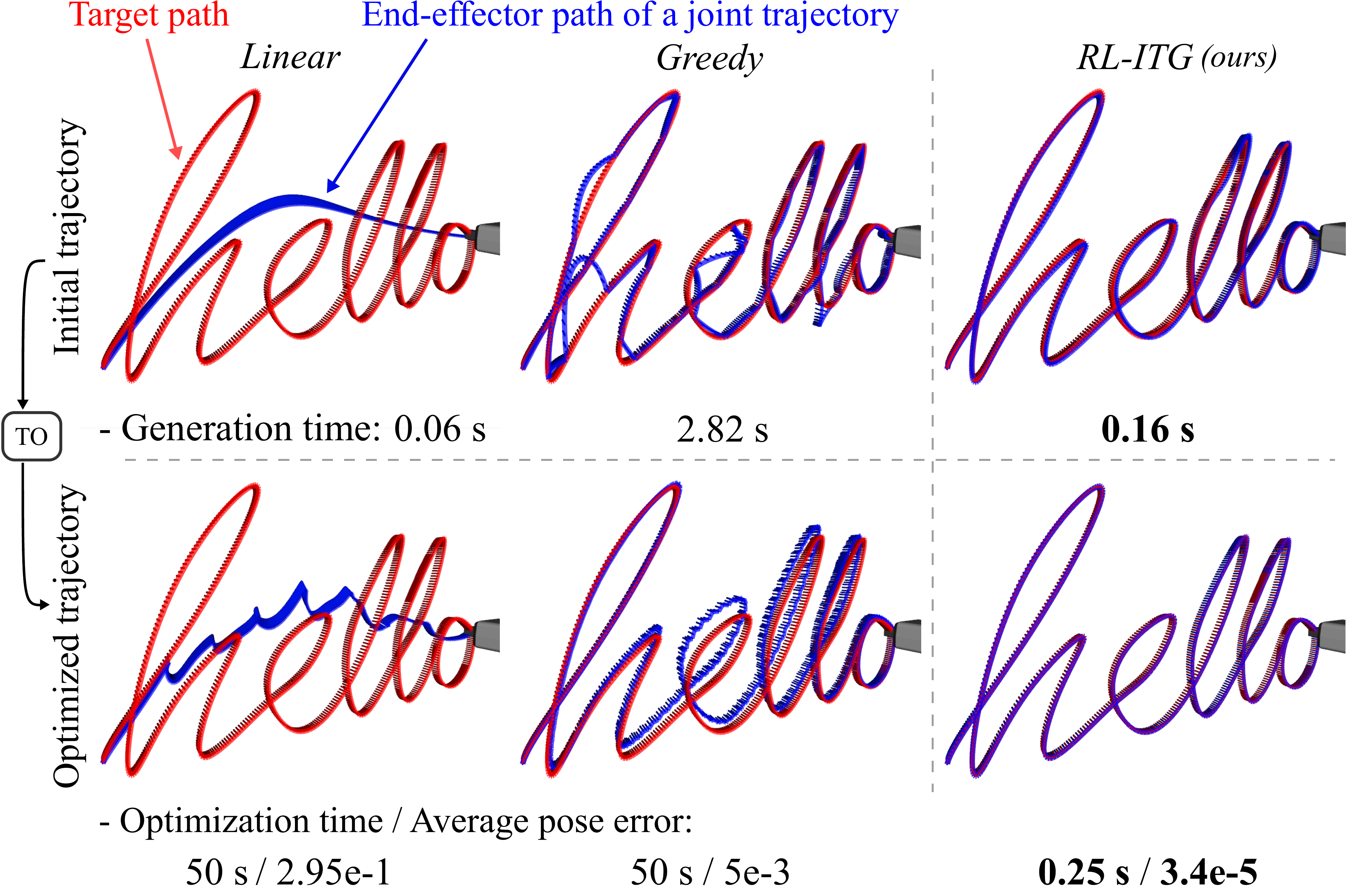}
    \vspace{-0.50cm}
	\caption{
        Given an exemplar path-following problem `Hello', first-row figures show initial trajectories of two baselines, \textit{Linear} \cite{schulman2013finding, zucker2013chomp} and \textit{Greedy} \cite{luo2012interactive, kang2020torm}, and ours, \textit{RL-ITG}.
        Putting these into a trajectory optimizer (TO), second-row figures show optimization results. 
        Ours quickly generates a high-quality initial trajectory rapidly converging to a better local optimum while the others get stuck in a boundary of a feasible solution.
        We return the solution of TO when the error is below 0.001 or the time limit of \SI{50}{\second} is exceeded.
	}
	\label{fig:main}
	\vspace{-0.65cm}
\end{figure}

To lower the complexity in searching for the solution, manipulation problems have adopted trajectory optimization (TO) methods that find a minimum-cost trajectory satisfying task-specific constraints~\cite{schulman2013finding, zucker2013chomp, luo2012interactive, kang2020torm, kalakrishnan2011stomp}.
However, the performance of converged trajectories and optimization time is sensitive to initial trajectories in non-convex problems~\cite{banerjee2020learning, melon2020reliable, ichnowski2020deep, melon2021receding}.
As initialization methods, researchers have often used 1) \textit{Linear}: linearly interpolating in configuration space~\cite{schulman2013finding, zucker2013chomp} or 2) \textit{Greedy}: greedily selecting IK solutions to maximally satisfy objectives~\cite{luo2012interactive, kang2020torm}. 
However, the limited consideration of the objectives and constraints leads the both approaches to sub-optimal or infeasible solutions for the path-following problem (see Fig.~\ref{fig:main}).
The global-search methods also could be leveraged for the initialization of TO. However, it still falls prey to poor local minima with a practical time limit since it is even slower than the \textit{Greedy}.
Therefore, we need an efficient way to rapidly generate high-quality initial trajectories to improve the performance of TO.

\textbf{Main Contributions}.
We present a learning-based method that fast synthesizes high-quality initial trajectories improving the TO performance for path-following problems of redundant manipulators (see Fig.~\ref{fig:main}).
Our method amortizes the extensive online computational load via example-guided reinforcement learning (RL)~\cite{peng2018deepmimic, RoboImitationPeng20} to offline learn a task-oriented motion satisfying the objectives and constraints. 
We call our method an RL-based initial trajectory generator (\textit{RL-ITG}). 
The RL scheme with our \textit{path-conditioned} Markov Decision Process (MDP) and large-scale training environments enables our method to learn a trajectory-generation policy applicable to various path-following setups.
In addition, we suggest null-space projected imitation reward inducing our policy to avoid collisions by learning kinematically feasible null-space motion captured in examples from an expert TO planner~\cite{kang2020torm}.

In experiments, we show that \textit{RL-ITG} helps TO methods significantly improve the solution's optimality, convergence speed (i.e., efficiency), and applicability over various problem setups compared to three baselines.
We also analyze which properties of the initial trajectory contribute to the performance improvement of TO.
Lastly, we verify the efficiency of our method via real-robot experiments with a seven-degree-of-freedom (DoF) mobile manipulator, a Fetch robot.

\begin{figure}[t]
	\centering
	\includegraphics[width=\columnwidth]{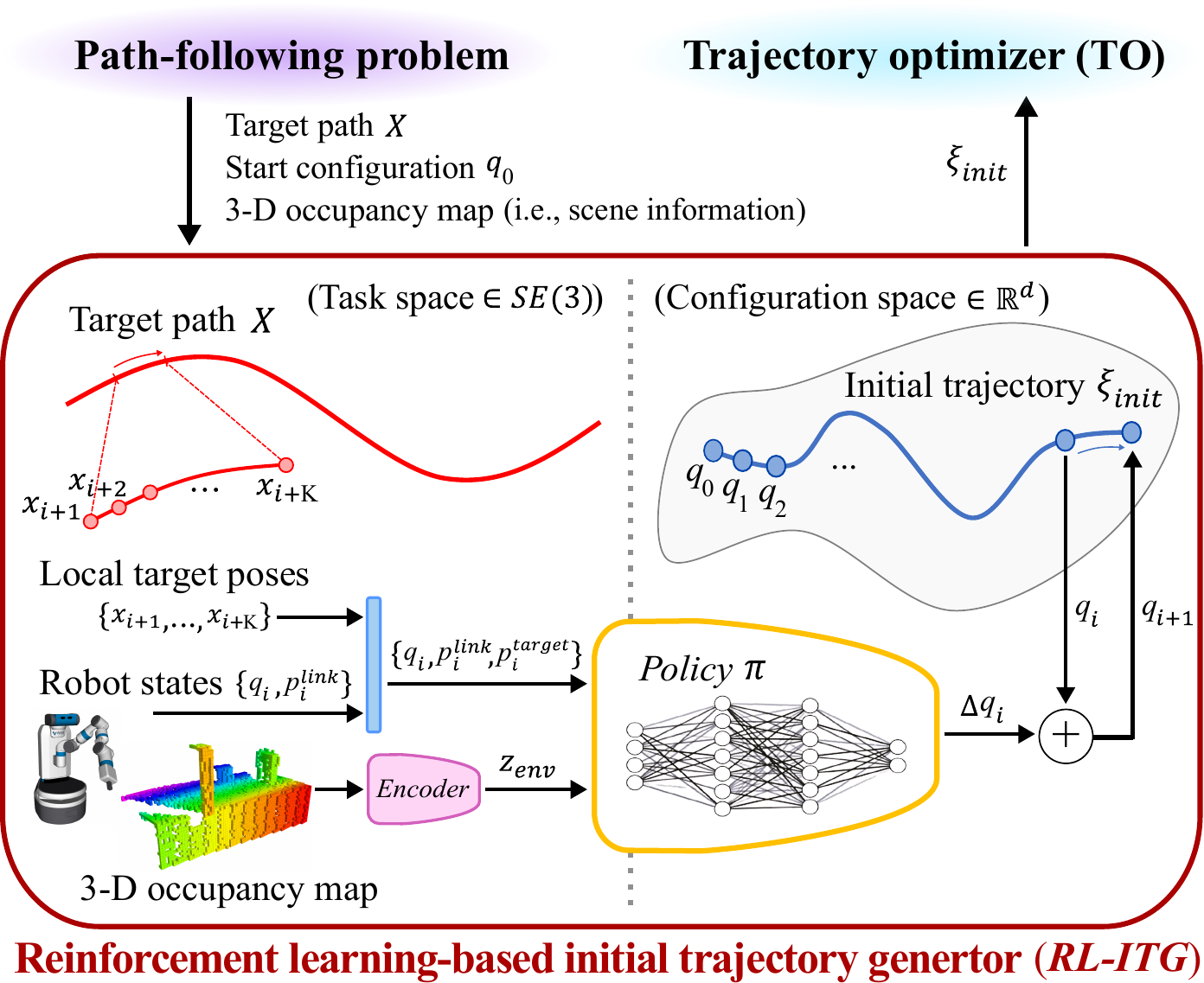}
    \vspace{-0.45cm}
	\caption{
        Illustration of how \textit{RL-ITG} produces an initial trajectory $\xi_{init}$.
        Starting from a start configuration $q_{0}$, our policy takes as input task-space information (i.e., local target poses, robot states, and a 3-dimensional occupancy map) and sequentially expands the trajectory in configuration space, taking into account objectives and constraints related to the path-following problem.
	}
	\label{fig:overall_framework}
	\vspace{-0.55cm}
\end{figure}

%% file: 3_1.tex
\section{Problem Definition and Motivation}
\label{sec:2}
Trajectory optimization for path-following problems is to transcribe 6-dimensional target paths into joint trajectories that satisfy various kinematic constraints such as mechanical limit, singularity, and collision. 
Researchers often represent the path as a sequence of poses $X=[x_0, x_1, ..., x_{N-1}] \in \mathcal{X}$, where $N$ is the number of poses in the path, $\mathcal{X}$ is the space of paths within a reachable workspace, and each pose is a pair of position and orientation; likewise, the joint trajectory is a sequence of joint configurations $\xi = [q_0, q_1, ..., q_{N-1}] \in \Xi$ evenly spaced in time with $\Delta t$, where $q \in \mathbb{R}^d$ is a configuration of a $d$ DoF manipulator and $\Xi$ is the Hilbert space of joint trajectories~\cite{rakita2019stampede, praveena2019user, kang2020torm}. 
Note that the redundant manipulator with a special Euclidean group $SE(3)$ has $d$ greater than 6.

There can be a number of kinematically feasible solution trajectories. 
To select the best one, TO requires setting an objective functional (i.e., cost function) $\mathcal{U} \in \mathbb{R}_{\geq 0}$, which often consists of multiple sub-objectives. 
For example, a state-of-the-art TO method targeted to the path-following problem~\cite{kang2020torm} defines a unified objective functional as
\begin{equation}
\mathcal{U}[\xi] = \mathcal{U}_{pose}[\xi] + \lambda_1 \cdot \mathcal{U}_{obs}[\xi] + \lambda_2 \cdot \mathcal{U}_{smooth}[\xi],
\label{eq:full_objective_f}
\end{equation}
where $\mathcal{U}_{pose}$, $\mathcal{U}_{obs}$, and $\mathcal{U}_{smooth}$ are an end-effector pose error\footnote{
    The pose error measures the distance between the end-effector path of the joint trajectory $\xi$ and the target path $X$. We use a weight of 0.17 for the rotational distance over the translational distance, as used in~\cite{holladay2019minimizing, kang2020torm}.
}, an obstacle avoidance cost, and a trajectory smoothness cost, respectively, regarding a trajectory $\xi$. 
$\lambda_1$ and $\lambda_2$ are constants. 
Further, the method considers kinematics constraints such as joint limits as well as constraints of singularity and collision avoidance. For more details, we refer the readers to \cite{kang2020torm}.

TO algorithms for the path-following problem are often sensitive to the quality of the initial trajectory $\xi_{init}$. 
Considering the objective functional $\mathcal{U}[\xi]$ represents the quality of the trajectory, we can design the problem of the trajectory initialization for TO as the minimization of $\mathcal{U}[\xi_{init}]$. 
Therefore, we aim to generate an initial trajectory having a small objective functional value or cost $\mathcal{U}[\xi_{init}]$ as fast as possible to help TO converge to a better optimal solution quickly.

%% file: 3_2.tex
\section{RL-based Initial Trajectory Generator}
\label{sec:3}
Our method, \textit{RL-ITG}, transforms target paths into joint trajectories by iteratively forward propagating our policy trained via the example-guided reinforcement learning (RL) (see Fig.~\ref{fig:overall_framework}).
Thus, our goal is to learn a policy that produces such high-quality, initial trajectories with a low-cost value of Eq.~(\ref{eq:full_objective_f}) to warm start TO. 
To this end, we first introduce how we formulate a Markov Decision Process (MDP) and an RL objective function, elucidate how we configure the training environments and the expert demonstration set, and then provide the training details concerning our policy.

\subsection{Formulation of an MDP and an RL objective function}
\label{sec:3-A}
Our RL formulation aims to learn a policy maximizing an RL objective function based on an MDP to generate high-quality, low-cost trajectories to be used as initial trajectories to TO.
To that end, we first formulate the path-following problem as a \textit{path-conditioned} MDP $\mathcal{M}_X = \langle \mathcal{S},\mathcal{A},\mathcal{R}_\mathcal{I},\mathcal{T}, \mathcal{Q}_0, \gamma \rangle_{X \sim \mathcal{X}}$, where $X$ is the target path, $\mathcal{S}$ is a state space, $\mathcal{A}$ is a action space, $\mathcal{R}_\mathcal{I}: \mathcal{S} \times \mathcal{A} \times \mathcal{I} \rightarrow \mathbb{R}$ is a time-varying reward function, where $\mathcal{I} = \{i\in\mathbb{N}_{0} | 0 \leq i < N\}$ is time steps, $\mathcal{T}: \mathcal{S} \times \mathcal{A} \rightarrow \mathcal{S}$ is a deterministic transition function, $\mathcal{Q}_0$ is a set of start configurations $q_0$, and $\gamma \in [0,1)$ is a discount factor. 
For $\mathcal{Q}_0$, we sample IK solutions at the first target pose $x_0 \in X$.
We then define a multi-path RL objective function to generalize the policy across the problems and obtain a unified policy:
\begin{equation}
\maximize_{\pi} \ \mathbb{E}_{ X \sim P(\mathcal{X})} \left[ 
            \mathbb{E}_{\substack{(s_i,a_i) \sim \rho_{\pi}\\q_0 \sim \mathcal{Q}_0}} \left[ 
                \sum_{i=0}^{N-1} \gamma ^{i} \cdot \mathcal{R}_i(s_i, a_i)   
                \right]   \right],
\label{eq:RL_objective_function}
\end{equation}
where $\rho_{\pi}$ is a trajectory distribution given a stochastic policy $\pi(a_i|s_i): \mathcal{S} \times \mathcal{A} \rightarrow \mathbb{R}_{\geq0}$ and the transition function $\mathcal{T}$~\cite{sutton2018reinforcement}, and $P(\mathcal{X})$ is a distribution of target paths.
Below, we describe how we define state and action spaces with reward functions.

\textbf{Scene-context-based state space}: 
We represent the state space $s_i$ at a time step $i$ as a tuple, $s_i = ( q_i, p_i^{link}, p_i^{target}, z_{env} )$: 
$p_i^{link} \in \mathbb{R}^{(d+1,9)}$ is a list of poses of the arm links. 
Note that we represent each pose in $SE(3)$ as a combination of a position vector ($\in \mathbb{R}^3$) and a 6-dimensional orientation vector ($\in \mathbb{R}^6$) to enhance the learning performance of the neural network with continuous state-space representation~\cite{zhou2019continuity}. 
$p_i^{target} \in \mathbb{R}^{(\mathrm{K},9)}$ is a list of relative distances from the current end-effector pose to local future target poses from time step $i+1$ to $i+\mathrm{K}$. 
Here, $\mathrm{K}$ is a hyper-parameter enabling the policy to take far-sighted, preemptive actions to secure a feasible action space in the configuration space.
$z_{env}$ is a scene-context vector embedding a perception information (i.e., a 3-dimensional (3-D) occupancy grid map) into a latent space using a pre-trained encoder (see Fig.~\ref{fig:overall_framework}).
This vector concisely represents the perception information, as used in~\cite{qureshi2020neural, ota2020deep}, and enables the policy to avoid collisions by optimizing the null-space motion of the redundant manipulator.
In this work, we represent all of the geometric states with respect to a base link frame.

\textbf{Action space}: We define the action space as a configuration difference $a_i = \Delta q_i \in \mathbb{R}^d$ relative to the current configuration $q_i$, and $q_{i+1} = q_i + \Delta q_i$ given the deterministic transition function $\mathcal{T}$.
Thus, the policy sequentially extends the trajectory for every step to complete the whole trajectory of length $N$.

\textbf{Example-guided path-following reward function}: 
We introduce a multi-objective guided reward function, adopting the example-guided RL, to induce the policy to minimize the cost function of TO (Eq.~(\ref{eq:full_objective_f})) and constraint violation rate:
\begin{equation}
\mathcal{R}_{i}= \mathcal{R}_{task,i} + \mathcal{R}_{im,i} +  \mathcal{R}_{cstr,i},
\end{equation}
where the reward terms on the right-hand side represent task, imitation, and constraint-related rewards at time step $i$, respectively. 
Note that, for the sake of simplicity, we omit the arguments in the reward functions. 

In detail, we form the task reward function $\mathcal{R}_{task,i}$ so that the end-effector path of the joint trajectory generated by our policy precisely matches the target path $X$.
$\mathcal{R}_{task,i}$ is composed of positional and rotational errors between the end-effector pose $\hat{x}_i(q_i)$ of the $i$-th configuration $q_i$ and the $i$-th target pose $x_i$ in the target path $X$.
Each pose is a tuple of position and quaternion orientation, $(x_i^{pos}, x_i^{quat})$. 
Then, we define the positional error $e^{pos}_i$ and the rotational error $e^{rot}_i$ given $q_i$ as
\begin{align}
e^{pos}_i &= || x_i^{pos} - \hat{x}_i^{pos}(q_i) ||_2, \\
e^{rot}_i &= 2 \cdot \cos^{-1}(|\langle x_i^{quat}, \hat{x}_i^{quat}(q_i) \rangle|),
\end{align}
where $\langle\cdot, \cdot\rangle$ is the inner product between two quaternions.
Then, instead of using the sum of negative errors as a combined reward, we normalize and reflect the relative importance of each error term $e$, similar to the parametric normalization \cite{rakita2018relaxedik}, defining a normalization function as
\begin{equation}
f(e, \mathbf{w}) = w_0 \cdot \exp(-w_1 \cdot e) - w_2 \cdot e^2 \quad \in \mathbb{R},
\end{equation}
where $\mathbf{w}=[w_0, w_1, w_2] \in \mathbb{R}_{\geq0}^{3}$ is a set of non-negative constants.
Thus, the task reward function is 
\begin{equation}
\mathcal{R}_{task,i}=f(e^{pos}_i,\mathbf{w}^{pos}) + f(e^{rot}_i,\mathbf{w}^{rot}) \cdot \mathds{1}_{\sqrt{e^{pos}_i} \leq \SI{5}{\cm}}, 
\end{equation}
where $\mathbf{w}^{pos}$ and $\mathbf{w}^{rot}$ are hyper-parameters of each error term. 
Further, we activate the rotational reward when the positional distance between the end-effector and target pose is within \SI{5}{\cm} to resolve the conflict between the two reward terms, as also mentioned in~\cite{allshire2021transferring}, using an indicator function $\mathds{1}$.

We also suggest the imitation reward $\mathcal{R}_{im,i}$ to make the policy learn the optimized null-space motion depicted in the demonstration set $\xi^{demo}$.
As redundant manipulators have a kinematic redundancy, many configurations could be possible to avoid collisions while following the target path.
We consider the case where the shortest distance between the robot arm and the external objects is maximized as the best configuration to reduce the possibility of collisions.
In addition, the selection of a consistent null-space motion on a constraint manifold is vital for generating a smooth trajectory without violating the kinematic constraints.
Therefore, to learn such motions, we prepare the demonstrations (i.e., examples) $\xi^{demo}$ optimized on a signed distance field (SDF)~\cite{zucker2013chomp} using the expert TO method~\cite{kang2020torm}.
Imitating examples enables efficient modeling of kinematically feasible null-space motion, but the non-optimality of the examples shapes a performance upper bound since the solutions from the TO without carefully curated initialization usually are poor local optima.
Thus, we present to project the imitation reward into the null space of the current configuration $q_i$ to resolve the performance degradation caused by naively mimicking sub-optimal data:
\begin{align}
\mathcal{R}_{im,i} &= f(e_{i}^{im}, \mathbf{w}^{im}), \\
e_{i}^{im} &= || (I-J(q_i)^{\dagger}J(q_i))(\xi^{demo}[i]-q_i) ||_2,
\end{align}
where $J(q_i)$ is the Jacobian matrix at the joint configuration $q_i$, $\dagger$ represents a Moore-Penrose inverse operation, $I$ is an identity matrix, and $\mathbf{w}^{im}$ is a set of non-negative constants.

Lastly, we define the constraint-related reward function $\mathcal{R}_{cstr,i}$ to penalize constraint violations regarding the collision, joint limits, and singularity and facilitate state exploration:
\begin{equation}
\mathcal{R}_{cstr,i} = \mathcal{R}_{C,i}+\mathcal{R}_{J,i}+\mathcal{R}_{S,i}+\mathcal{R}_{E,i},
\end{equation}
where $\mathcal{R}_{C,i} = -10 \cdot \mathds{1}_{collision}$, $\mathcal{R}_{J,i} = -1 \cdot \mathds{1}_{q_i < q_{min} \bigcup q_i > q_{max}}$, and $\mathcal{R}_{S,i} = -0.1 \cdot \mathds{1}_{\det(J(q_i)J(q_i)^{T}) < 0.005}$. 
We also terminate training episodes with an early termination reward when the end-effector position is more than \SI{20}{\cm} away from the $i$-th target position, $\mathcal{R}_{E,i}= -3 \cdot \mathds{1}_{\sqrt{e^{pos}_i} > \SI{20}{\cm}}$, to enhance the training efficiency of state-space exploration, as proposed in \cite{peng2018deepmimic}.

\begin{figure}[t!]
    \centering
    \includegraphics[width=\columnwidth]{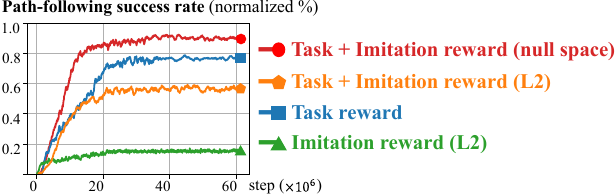}
    \vspace{-0.5cm}
    \caption{
        Learning curves of four reward function compositions.
        We measure the success rate on 20 evaluation sets randomly generated with the procedure in Sec.~\ref{sec:3-B} at every $10^4$ steps and consider one experiment successful when distances between the end effector pose and the target pose are within \SI{5}{\cm} positionally and \SI{3}{\degree} rotationally without any collision at all time steps.
    }
    \vspace{-0.55cm}
    \label{fig:reward_comapre}
\end{figure}

Fig.~\ref{fig:reward_comapre} illustrates the benefits of combining our null-space projected imitation reward $\mathcal{R}_{im,i}$ with task reward $\mathcal{R}_{task,i}$ over other reward compositions, with the constraint-related reward $\mathcal{R}_{cstr,i}$ used in common.
Here, we name the reward naively mimicking sub-optimal examples with the error term $e_{i}^{im, L_2} = ||\xi^{demo}[i]-q_i ||_2$ as an $L_2$ imitation reward $\mathcal{R}^{L_2}_{im,i}$.
Our reward formulation (a red line in Fig.~\ref{fig:reward_comapre}) surpassed other reward settings in training performance and efficiency thanks to the kinematically feasible null-space motion guidance of examples by settling the performance deterioration caused by examples' sub-optimality.
On the other hand, an orange line shows performance degradation when the task reward (a blue line) is combined with the simple $L_2$ imitation reward $\mathcal{R}^{L_2}_{im,i}$.

\subsection{Generation of training environments and examples}
\label{sec:3-B}
We construct various path-following problems as training environments to help our policy applicable to the different problem setups (i.e., target path, external objects, and start configuration).
For the set of target paths $X$, we first collect 5,000 \textit{valid} end-effector poses within $x:\interval{0.2}{1.2} \times y:\interval{-0.7}{0.7} \times z:\interval{0.0}{1.2}$ (unit: \si{\m}) and an unlimited orientation range.
We set the positional range considering the reachable workspace of the Fetch robot.
We consider the pose \textit{valid} when at least one IK solution satisfies the collision-free constraint.
We then randomly sample 5 to 8 \textit{valid} poses as waypoints and interpolate them at \SI{0.5}{\cm} positional intervals along a B-spline curve with orientational interpolation using a spherical linear interpolation.
We lastly filter out the path if any intermediate pose $\{x_i \in X \,|\, i \in [0,N-1] \}$ is not \textit{valid}.
We thus collect 5,000 paths without external obstacles.
To consider the obstacles, we generate 500 random scenes in the form of tabletops with different sizes of tables and objects, and we collect 20 paths per scene with the same procedure described above.
Thus, we collect 5,000 paths without obstacles and 10,000 paths with obstacles (see examples in a pink box in Fig.~\ref{fig:test_scenes}).
In total, we collect 30,000 training problems by randomly sampling two IK solutions for $q_0$ at the first target pose $x_0$ per path $X$.

In addition, we collect examples, optimized joint trajectories $\xi^{demo}$, on the 30,000 problems using the expert TO method, trajectory optimization of a redundant manipulator (TORM)~\cite{kang2020torm}, with a \SI{120}{\second} time budget.
We gathered the data in parallel. This amounts to 1,000 hours for serial collection.

\subsection{Training details}
\label{sec:3-C}
We make use of a soft actor-critic framework \cite{haarnoja2018soft} to train our policy maximizing Eq.~(\ref{eq:RL_objective_function}). 
We compose the policy and double Q networks consisting of three fully-connected layers with 1,024 neurons. 
We refer to their learning parameters as $\theta$.
We model the policy as a stochastic diagonal Gaussian policy and bound the output range into $\interval{-0.26}{0.26}$ (unit: \si{\radian}) considering velocity limits to enforce that the generated trajectory naturally satisfies connectivity and smoothness: $\pi_\theta(a|s) \sim 0.26 \cdot \tanh(\mathcal{N}(\mu_\theta(s), \Sigma_\theta(s)))$.
For the hyper-parameters of the MDP, we empirically find the best performing values as $\mathrm{K}=6$, $dim(z_{env})=32$, $\mathbf{w}^{pos}=[2, 65, 30]$, $\mathbf{w}^{rot}=[2, 5, 0]$, and $\mathbf{w}^{im}=[1,15,0.5]$.
Training of the policy required \num{3e+7} simulation steps in total, which took about 144 hours on the standard desktop equipped with an Intel i9-9900K and an RTX 2080 TI.

We adopt a variational auto-encoder (VAE)~\cite{kingma2013auto} framework to generate the scene-context vector $z_{env}$.
As training data, we collected 5,000 random scenes configured in the form of tables with objects scattered on the table.
We represented the scene information as the 3-D occupancy grid map in range of $x:\interval{0.2}{1.2} \times y:\interval{-0.7}{0.7} \times z:\interval{0.2}{1.2}$ (unit: \si{\m}) with \SI{2}{\cm} resolution using SuperRay~\cite{kwon2016super} with an RGB-D camera, Primesense Carmine 1.09. 
The training took about 3 hours for 500 epochs on the same optimizer and machine. 
For the rest of the configurations, please refer to the code\footnote{https://github.com/MinsungYoon/RL-ITG}.

%% file: 4.tex
\section{Experiments}
The major advantage of \textit{RL-ITG} is the competency in quickly generating high-quality trajectories with a low cost and constraint violation rate. 
Thus, we expect this advantage leads to the improved performance of TO by warm starting TO.
In experiments, we aim to corroborate the expectation by comparing \textit{RL-ITG} with three other baseline methods.
We also look into which facets of initial trajectories contribute to performance improvement.
Lastly, real-world experiments show how \textit{RL-ITG} efficiently solves path-following problems.

{
    \begin{figure}[t!]
    	\centering
    	\includegraphics[width=\columnwidth]{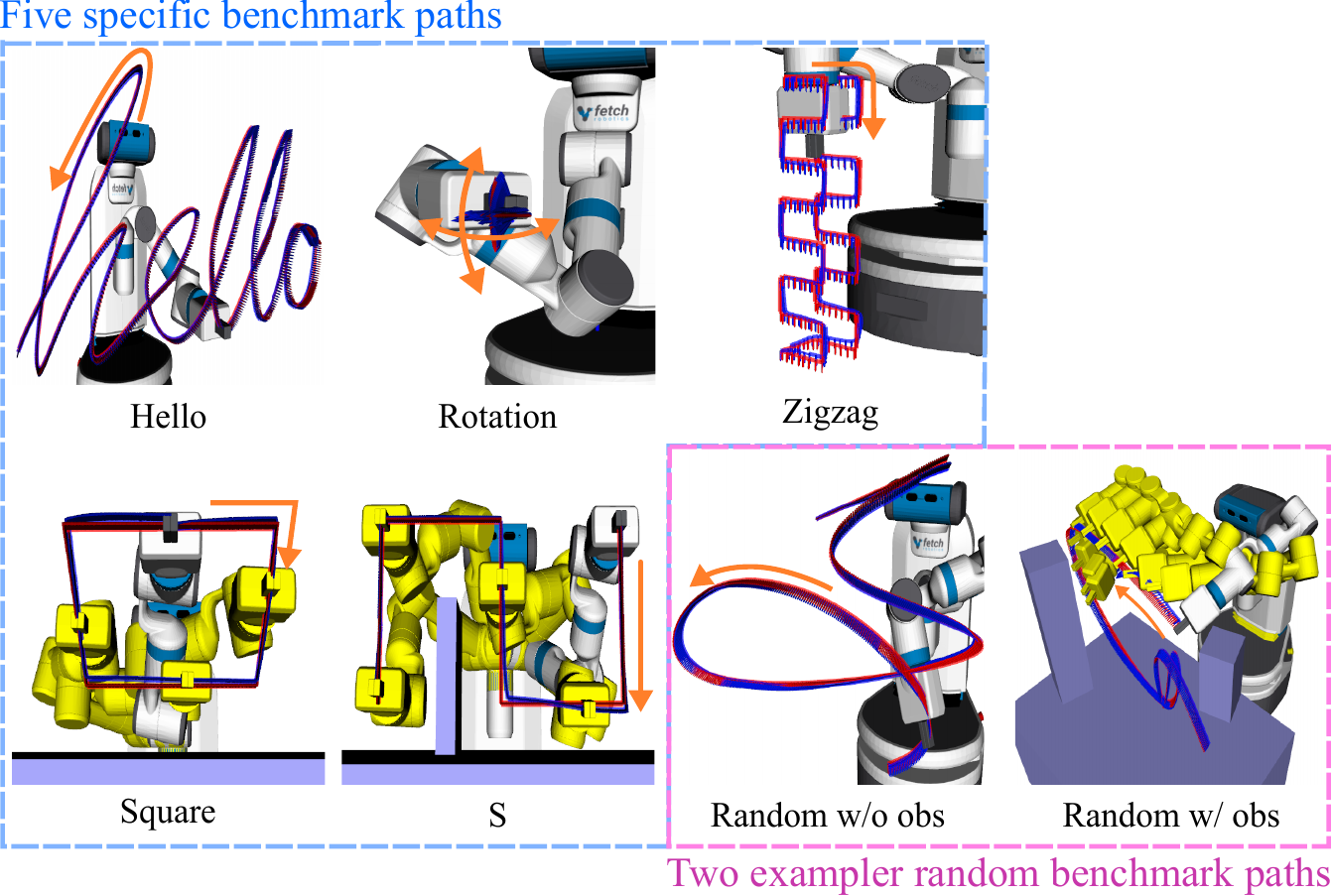}
    	\vspace{-0.43cm}
    	\caption{
        		Visualization of five specific and two exemplar random target paths (red lines) used in evaluations.
                Orange arrows indicate the progress direction of the paths.
                Blue lines are the end-effector paths of the initial joint trajectories generated by our method, \textit{RL-ITG}.
                In `Square', `S', and `Random w/ obs', the original color of the robot represents the start configuration, and the yellow trail (i.e., some intermediate configurations) shows that the trajectories avoid collisions with external objects and the robot body via optimizing the null-space motion while precisely following the paths.  
    	}
    	\label{fig:test_scenes}
    	\vspace{-0.6cm}
    \end{figure}
}

\subsection{Experimental setup}
\label{sec:4-A}

We first prepared two representative and one learning-based trajectory initialization methods as our baselines:
\begin{itemize}[leftmargin=*]
    \item \textbf{\textit{Linear}}:
    \textit{Linear} returns a linearly interpolated trajectory in the configuration space.
    Considering that the goal configuration is not given in the target path, we selected an IK solution having a minimum $L_2$ distance with the start configuration $q_0$ or collision-free one at the last pose $x_{N-1}$.
    \item \textbf{\textit{Greedy}}~\cite{kang2020torm}:
    \textit{Greedy} extracts the sub-sampled poses from $X$ with $10$ intervals. 
    Then, starting from the start configuration $q_0$, this finds $100$ random IK solutions at each next sub-sampled pose and linearly interpolates with the best IK solution minimizing the TO objective function (Eq.~(\ref{eq:full_objective_f})). 
    We used the same hyper-parameter values as used in~\cite{kang2020torm}.
    \item \textbf{\textit{BC-ITG}}: 
    We prepared \textit{BC-ITG} as a baseline in perspective of the learning method.
    We trained the neural network with a behavior-cloning (BC) framework~\cite{bojarski2016end}, which mimics the demonstration set $\xi^{demo}$ with a mean squared error loss.
\end{itemize}

We also demonstrated the applicability of each initialization method for the two types of TO: TORM~\cite{kang2020torm} and Trajectory Optimization for Motion Planning (TrajOpt)~\cite{schulman2013finding}.
We set TORM, a first-order optimization approach, to iteratively optimize and explore new initial trajectories to avoid bad local minima within \SI{50}{\second} time budget. 
On the other hand, TrajOpt's update is performed using a quadratic solver. Thus, one iteration takes from \SI{3}{\second} to \SI{14}{\second}, so we made the trajectory fully converge within \SI{150}{\second} without exploring new trajectories.

We used five specific target paths $X$ as our benchmark paths (see Fig.~\ref{fig:test_scenes}). 
Three paths (`Hello', `Rotation', and `Zigzag') are without external obstacles, and two paths (`Square' and `S') are with external obstacles. 
In the case of `Hello', `Zigzag', `Square', and `S', we fixed the orientation on the path. 
On the other hand, in the `Rotation', we fixed the position on the path while varying the orientation in the range of $\pm$\SI{45}{\degree} along the direction of pitch and yaw axes in order. 
We also prepared a `Random' benchmark set following the same procedure in Sec.~\ref{sec:3-B} to compare and evaluate the performance of the two representative methods, \textit{Linear} and \textit{Greedy}, on the randomly generated problems with ours, \textit{RL-ITG}.
`Random' consists of 100 paths without external objects and 1,000 paths with objects.
Then, we sampled 100 and 5 collision-free start configurations $q_0$ per specific and random benchmark path $X$, respectively, to obtain reliable statistical results since the difficulty of path-following problems largely depends on the start configurations~\cite{praveena2019user}.
In total, we collected 6,000 benchmark problems for evaluation.
The length of each benchmark path follows $N_{Hello}=553$, $N_{Rotation}=209$, $N_{Zigzag}=227$, $N_{Square}=320$, and $N_{S}=301$. 
The statistical distribution of the `Random' paths is $N_{Random} \sim \mathcal{N}(626,120)$.

Lastly, we set the transition time $\Delta t$ of the joint trajectory $\xi$ as \SI{0.1}{\second} to check for the velocity limit constraint and objective function's coefficients as $\lambda_1=10$ and $\lambda_2= \frac{5}{N+1}$ to adjust the scale between each sub-objective function as used in~\cite{kang2020torm}.

%% file: 5.tex
\subsection{Experimental results}
\label{sec:5}

\begin{figure}[t]
    \centering
    \includegraphics[width=\columnwidth]{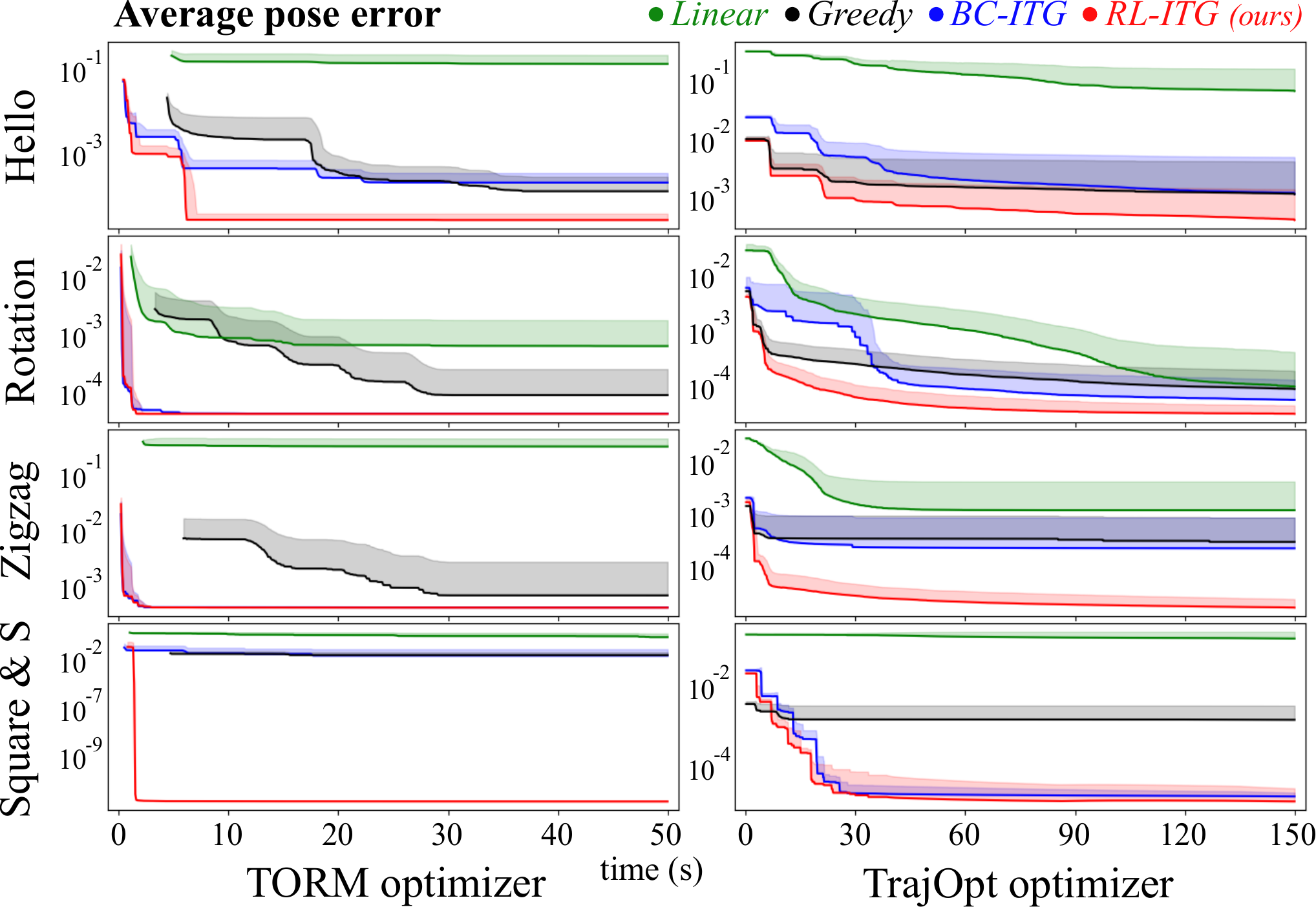}
    \vspace{-0.53cm}
    \caption{
        Comparison of average pose error convergence of four initialization methods over eight combination experiments of four benchmark sets (i.e., \textit{Hello}, \textit{Rotation}, \textit{Zigzag}, and \textit{Square \& S}) and two optimization methods (i.e., TORM~\cite{kang2020torm} and TrajOpt~\cite{schulman2013finding}). 
        For each sub-figure, the y-axis is the pose error in the log scale, the x-axis is the elapsed time (s), and the shaded area represents one standard deviation over the mean.
    }
    \vspace{-0.2cm}
    \label{fig:opt_progress}
\end{figure}

We first show the effect of our method, \textit{RL-ITG}, on the optimization performance of TO by combining each initialization method with two different trajectory optimization approaches, TORM and TrajOpt.
Fig.~\ref{fig:opt_progress} shows the convergence speed and optimality of the average pose error over the optimization time.
\textit{RL-ITG} exhibits superior convergence performance on all benchmark sets even though the initial error is often higher than \textit{Greedy}.
These results imply that, in terms of initialization, not only is the end-effector matching objective important, but also the other aspects (e.g., continuity of the initial trajectories on a constraint manifold) are crucial for convergence to better local optima via TO.
In addition, the relatively high initial pose error of \textit{RL-ITG} indicates that integrating learning and planning is an important strategy that works in a complementary manner.
The pose error of \textit{Greedy} with TORM drops step-wise, and the convergence time takes a long time since it searches the initial trajectories through repeated restarts.
On the other hand, the convergence performance of \textit{BC-ITG} is fairly competitive with \textit{RL-ITG}, but it lacks applicability for the different benchmarks with the finite demonstration set $\xi^{demo}$ we prepared.

We also measured success rates on a set of benchmarks to evaluate our method's ability to solve various path-following problem configurations, as shown in Table~\ref{table:opt_suc_and_feasible_rate}.
We consider the optimized solution successful when the trajectory returned after the time budget satisfies the constraints, and the positional and rotational errors are smaller than each threshold value in the parentheses.
We set the threshold values differently since the scale of the converged error values is different for each problem.
\textit{RL-ITG} improves the performance of TO methods by consistently maintaining the highest success rate in all benchmark sets.
In particular, \textit{RL-ITG} shows a noticeable performance improvement compared to \textit{Linear} and \textit{Greedy} on the `Random' benchmarks, which are composed to evaluate the applicability of the basic initialization methods to the randomly generated problem setups.
Note that we disabled the iterative restarts of \textit{Greedy} with TORM for a fair comparison in this experiment. 

{
    \begin{table}[t]
    \centering
    \begin{tabular}{|c|c|S[table-format = 3.1, detect-weight, detect-shape, detect-mode]|S[table-format = 3.1, detect-weight, detect-shape, detect-mode]|S[table-format = 3.1, detect-weight, detect-shape, detect-mode]|S[table-format = 3.1, detect-weight, detect-shape, detect-mode]|}
    \hline
    \multirow{2}{*}{Benchmark set} &\multirow{2}{*}{TO} &\multicolumn{4}{c|}{Method} \\ \cline{3-6}
     & & \textit{Linear}   &  \textit{Greedy}      &  \textit{BC-ITG} &  \textit{RL-ITG} \\ \hline\hline
    
    \multirow{2}{*}{\specialcell{Hello\\{\scriptsize (\SI{0.1}{\cm}, \SI{0.1}{\degree})} }}    
        & TORM  &  32.0        &  68.0          &  62.0     &  \bftabnum 87.0     \\  
        & TrajOpt   &  46.0      &  99.0      & 92.0     &  \bftabnum 100.0 \\ \hline\hline
                
    \multirow{2}{*}{\specialcell{Rotation\\{\scriptsize (\SI{0.1}{\cm}, \SI{0.1}{\degree})} }}    
        & TORM  &  93.0        &  98.0         &  87.0     &  \bftabnum 99.0   \\  
        & TrajOpt   &  99.0      & \bftabnum 100.0      & \bftabnum 100.0     &  \bftabnum 100.0 \\ \hline\hline
        
    \multirow{2}{*}{\specialcell{Zigzag\\{\scriptsize (\SI{1}{\cm}, \SI{1}{\degree})} }}    
        & TORM  &  3.0      &  74.0      &  94.0    &  \bftabnum 98.0   \\  
        & TrajOpt   &  73.0      &  77.0      & 87.0     &  \bftabnum 97.0 \\ \hline\hline
        
    \multirow{2}{*}{\specialcell{Square \& S\\{\scriptsize (\SI{1}{\cm}, \SI{1}{\degree})} }}    
        & TORM  &  6.0        &  89.0        &  76.0      &  \bftabnum 100.0   \\  
        & TrajOpt   &  22.0      &  \bftabnum 100.0      & \bftabnum 100.0     &  \bftabnum 100.0 \\ \hline\hline
        
    \multirow{2}{*}{\specialcell{Random\\{\scriptsize (\SI{1}{\cm}, \SI{1}{\degree})} }}    
        & TORM  &  2.3     &  19.1       &  61.6       &  \bftabnum 88.9   \\  
        & TrajOpt   &  6.5      &  90.3      & 85.1     &  \bftabnum 99.3 \\ \hline
    \end{tabular}
    \caption{
        Comparison of two extended trajectory-optimization approaches with each trajectory initialization method in terms of success rate (\si{\%}). We consider an optimized trajectory successful if the trajectory satisfies kinematic feasibility constraints and the average positional and rotational errors are lower than certain threshold values shown in the parentheses.
    }
    \label{table:opt_suc_and_feasible_rate}
    \vspace{-0.65cm}
    \end{table}
}

{
    \begin{figure*}[t!]
    	\centering
    	\includegraphics[width=2\columnwidth]{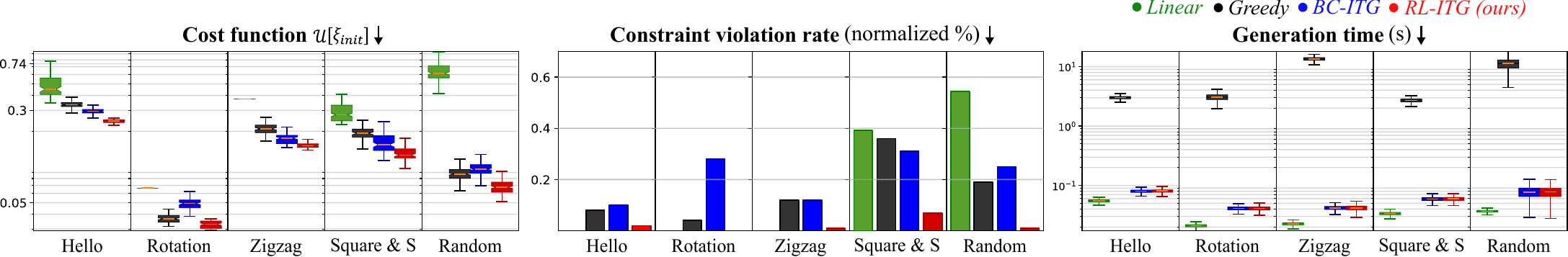}
        \caption{
        		Comparative analysis of the four initial-trajectory generation methods in five types of simulated benchmarks. The x- and y-axes are the type of benchmark problems and the performance metrics, respectively.
        }
        \label{fig:Initial_trajectory_quantitative_analysis}
        \vspace{-0.5cm}
    \end{figure*}
}

We then analyze which aspects of the initial trajectory are responsible for finding better optimal solutions via TO by comparing each initialization method quantitatively and qualitatively.
Fig.~\ref{fig:Initial_trajectory_quantitative_analysis} shows the quantitative results in terms of three metrics: 1) objective functional value of TO $\mathcal{U}[\xi_{init}]$ (Eq.~(\ref{eq:full_objective_f})), 2) constraint violation rate considering the collision-free and velocity-limit constraints, and 3) generation time.
\textit{RL-ITG} shows the lowest objective functional value in all benchmark sets.
In addition, \textit{RL-ITG} shows a lower constraint violation rate than \textit{Greedy} and \textit{BC-ITG}.
\textit{Linear} does not consider the target path and collisions with external objects.
Thus, in the absence of environmental obstacles, there is little chance of violating the constraints.
On the other hand, it shows the highest constraint violation rate in the benchmarks considering obstacles.
In terms of computational efficiency (i.e., generation time), the learning-based methods, \textit{BC-ITG} and \textit{RL-ITG}, naturally outperformed the IK-based method, \textit{Greedy}.
The learning-based methods quickly generate smooth trajectories by forward-propagating the network even in novel environments by taking the path and perception information as input.
Fig.~\ref{fig:test_scenes} qualitatively exhibits the initial trajectories synthesized by \textit{RL-ITG} on the exemplar benchmarks.

Fig.~\ref{fig:Initial_trajectory_quality} and \ref{fig:manifold} further examine each method's output on specific examples.
In Fig.~\ref{fig:Initial_trajectory_quality}, \textit{Greedy} has a relatively small pose error but shows the worst smoothness (i.e., trajectory length) due to a practical time limit for exploring IK solutions.
\textit{RL-ITG} shows competitive precision on the pose error with \textit{Greedy} while getting balanced between the sub-objectives (i.e., pose error and smoothness).
Fig.~\ref{fig:manifold-a} shows initial trajectories and a constraint manifold regarding one `Square' benchmark.
\textit{Greedy} violated the constraints by suddenly jumping on the manifold, and it chose the mode leading to a lengthy trajectory.
\textit{RL-ITG} contrarily selected a mode that creates a short-length trajectory and maintained consistent proximity to the constraint manifold.
Fig.~\ref{fig:manifold-b} shows snapshots in the task space when \textit{Greedy} momentarily jumps on the manifold compared with the smooth transition of \textit{RL-ITG}.

{
    \begin{figure}[t]
        \centering
        \includegraphics[width=\columnwidth]{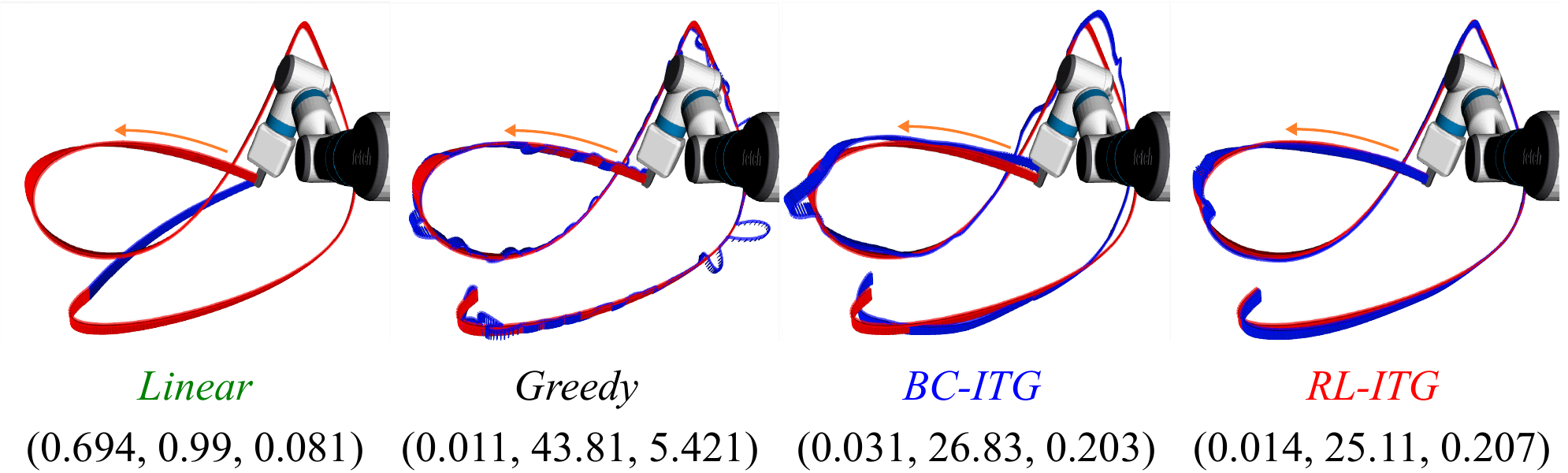}
        \vspace{-0.4cm}
        \caption{
            Comparison of initial trajectories on one of the `Random' problems.
            The red and blue lines are the target and end-effector paths from generated initial trajectories. 
            Numbers within the parenthesis represent the average pose error, the trajectory length (\si{\radian}), and the generation time (\si{\second}) in order.
            The orange arrow and the robot posture indicate the target path direction and the start configuration, respectively.
        }
        \label{fig:Initial_trajectory_quality}
        \vspace{-0.6cm}
    \end{figure}
}

{
    \begin{figure}[t]
        \vspace{-0.3cm}
        \centering 
        \subfigure[Constraint manifold with initial trajectories]{
            \includegraphics[width=0.65\columnwidth]{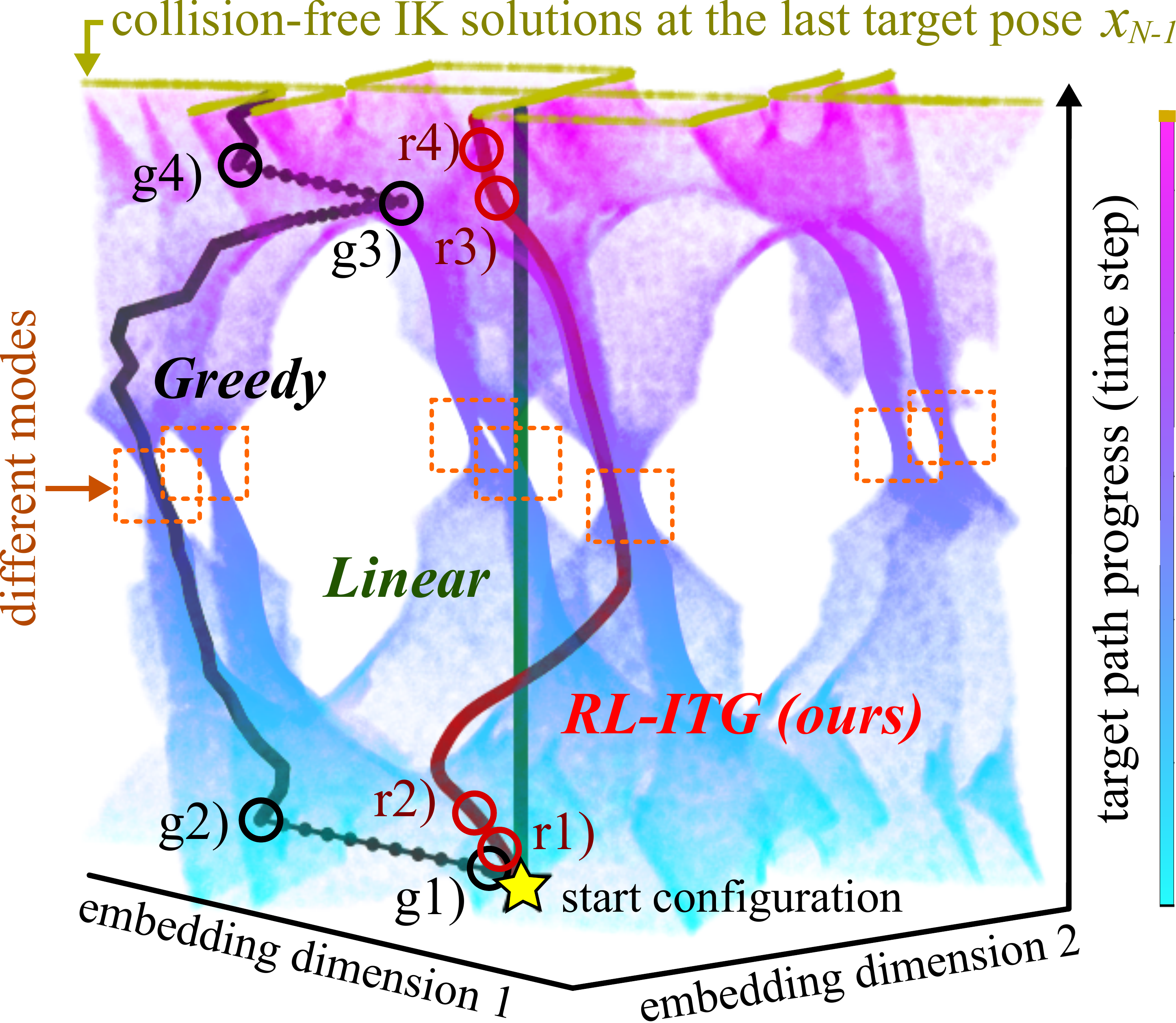}
            \label{fig:manifold-a}
        }
        \subfigure[Snapshots of \textit{Greedy} and \textit{RL-ITG} trajectories]{
            \includegraphics[width=0.84\columnwidth]{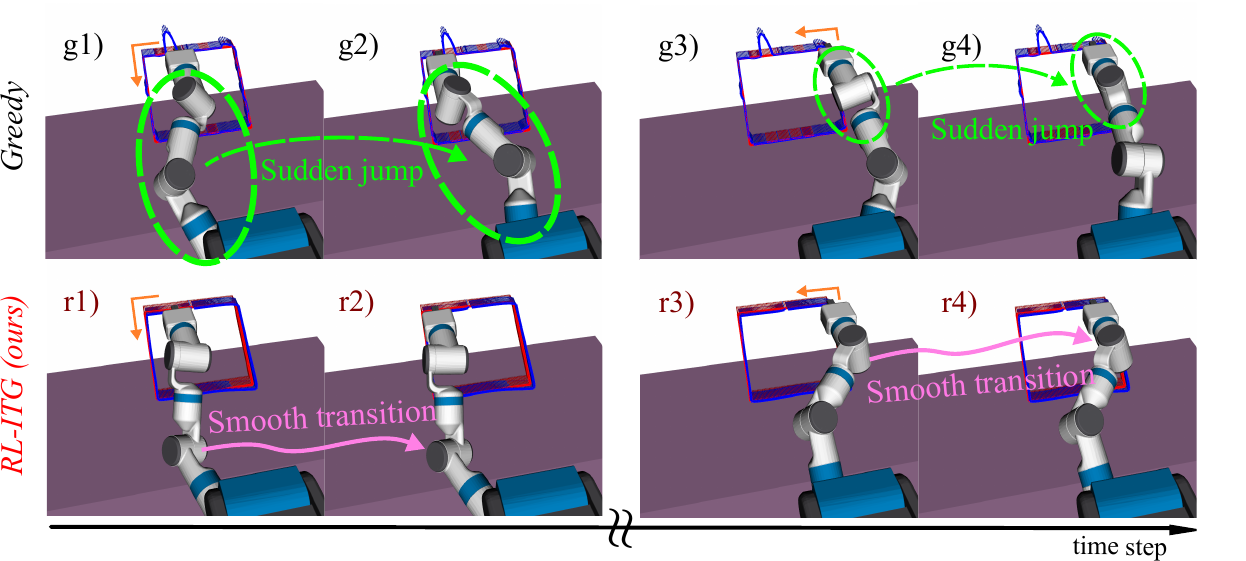}
            \label{fig:manifold-b}
        }
        \vspace{-0.2cm}
        \caption{
        (a) A visualization of initial trajectories overlaid on the projected constraint manifold of the `Square' benchmark. 
        To approximate the constraint manifold, we sampled 5,000 collision-free IK solutions per pose along the target path and then stacked them along the z-axis after reducing configuration dimensionality using principal component analysis (PCA)~\cite{abdi2010principal}.
        (b) Snapshots of the \textit{Greedy} and \textit{RL-ITG} trajectories at the moment marked in (a).
        } 
        \label{fig:manifold}
        \vspace{-0.55cm}
    \end{figure}
}

We finally verified the efficiency of \textit{RL-ITG} on the real robot compared with \textit{Greedy}.
With the experiments on the exemplar benchmarks of `Hello', `Rotation', `Zigzag', `Square', `S', and `Random w/o obs', \textit{RL-ITG} boosted the total execution time 1.2, 1.13, 1.25, 1.9, 1.6, and 2.2 times faster than \textit{Greedy}, respectively.
We turned on a timer when the system started up and executed the solution trajectory, which is returned with the same criteria in Table~\ref{table:opt_suc_and_feasible_rate}.
The result shows that the optimized trajectories with our initialization method, \textit{RL-ITG}, have shorter trajectory lengths with a fast solution generation time.
We have attached videos of each real-world experiment in the supplementary video.

\section{Conclusion}
We have presented \textit{RL-ITG} method that rapidly finds high-quality initial trajectories for the path-following problems of the redundant manipulator.
The simulated and real-world experiments have shown that \textit{RL-ITG} improves the optimality, efficiency, and applicability of TO methods with high-quality initial guesses.
This work focused on offline planning for long-horizon path-following problems in a static environment, confining the scene configuration in the form of a tabletop.
In future work, we plan to diversify the scenes and integrate our method with the receding horizon control scheme to handle dynamic environments and time-varying target paths.